\pdfoutput=1

\documentclass[11pt]{article}

\usepackage[final]{acl}

\usepackage{times}
\usepackage{latexsym}
\usepackage{amsmath}
\usepackage{amssymb}
\usepackage{mathtools}
\usepackage{amsthm}
\usepackage{multirow}
\usepackage{array}
\usepackage{cancel}
\usepackage{booktabs}
\usepackage{multirow}
\usepackage{bigstrut}
\usepackage{scalerel}
\usepackage{cleveref}
\usepackage{hyperref}
\usepackage{longtable}

\DeclareMathSymbol{\minus}{\mathbin}{AMSa}{"39}
\newcommand{\exper}[1]{\textsc{#1}}

\usepackage[T1]{fontenc}

\usepackage[utf8]{inputenc}

\usepackage{microtype}

\usepackage{inconsolata}

\usepackage{pgfplots}
\pgfplotsset{compat=1.18}
\usepackage{tikz}
\usetikzlibrary{arrows.meta}

\title{An Information Bottleneck Perspective for Effective Noise Filtering on Retrieval-Augmented Generation}

\author{Kun Zhu$^{1}$\footnotemark[1], Xiaocheng Feng$^{1,2}$\thanks{Equal Contribution}, Xiyuan Du$^{1}$, Yuxuan Gu$^{1}$, Weijiang Yu$^{3}$,\\
{\bf Haotian Wang$^{1}$}, {\bf Qianglong Chen$^{4}$}, {\bf Zheng Chu$^{1}$}, {\bf Jingchang Chen$^{1}$}, {\bf Bing Qin$^{1,2}$}\thanks{Corresponding Author}\\
  $^{1}$Harbin Institute of Technology\quad \quad  $^{2}$ Peng Cheng Laboratory\\
$^{3}$ Sun Yat-sen University \quad \quad \quad  $^{4}$  	Zhejiang University\\
  \texttt{\{kzhu,xcfeng,xydu,yxgu,zchu,jcchen,qinb\footnotemark[2]\}@ir.hit.edu.cn} \\
  \texttt{\{weijiangyu8, wanght1998, chenqianglong.ai\}@gmail.com}}

\begin{document}
\maketitle
\begin{abstract}
Retrieval-augmented generation integrates the capabilities of large language models with relevant information retrieved from an extensive corpus, yet encounters challenges when confronted with real-world noisy data.
One recent solution is to train a filter module to find relevant content but only achieve suboptimal noise compression.
In this paper, we propose to introduce the information bottleneck theory into retrieval-augmented generation.
Our approach involves the filtration of noise by simultaneously maximizing the mutual information between compression and ground output, while minimizing the mutual information between compression and retrieved passage.
In addition, we derive the formula of information bottleneck to facilitate its application in novel comprehensive evaluations, the selection of supervised fine-tuning data, and the construction of reinforcement learning rewards.
Experimental results demonstrate that our approach achieves significant improvements across various question answering datasets, not only in terms of the correctness of answer generation but also in the conciseness with $2.5\%$ compression rate. 



\end{abstract}

\section{Introduction}

Large language models represent a significant advancement in natural language understanding and generation, with the capability to process and produce human-like language at an unprecedented scale and complexity \cite{achiam2023gpt, touvron2023llama, team2023gemini}. 
Nonetheless, large language models have several drawbacks, such as hallucination \cite{huang2023survey} and lacking knowledge for specific domains or highly specialized queries \cite{kandpal2023large}.
Retrieval-augmented generation \cite{lewis2020retrieval} has gained attention for its ability to incorporate information from external knowledge sources during the inference stage. 
By combining retrieval-based methods with generative models, this approach can improve the relevance, coherence, and factual accuracy of text generation \cite{gao2023retrieval}. 

\begin{figure}[t]
    \centering
    \includegraphics[width=1\columnwidth]{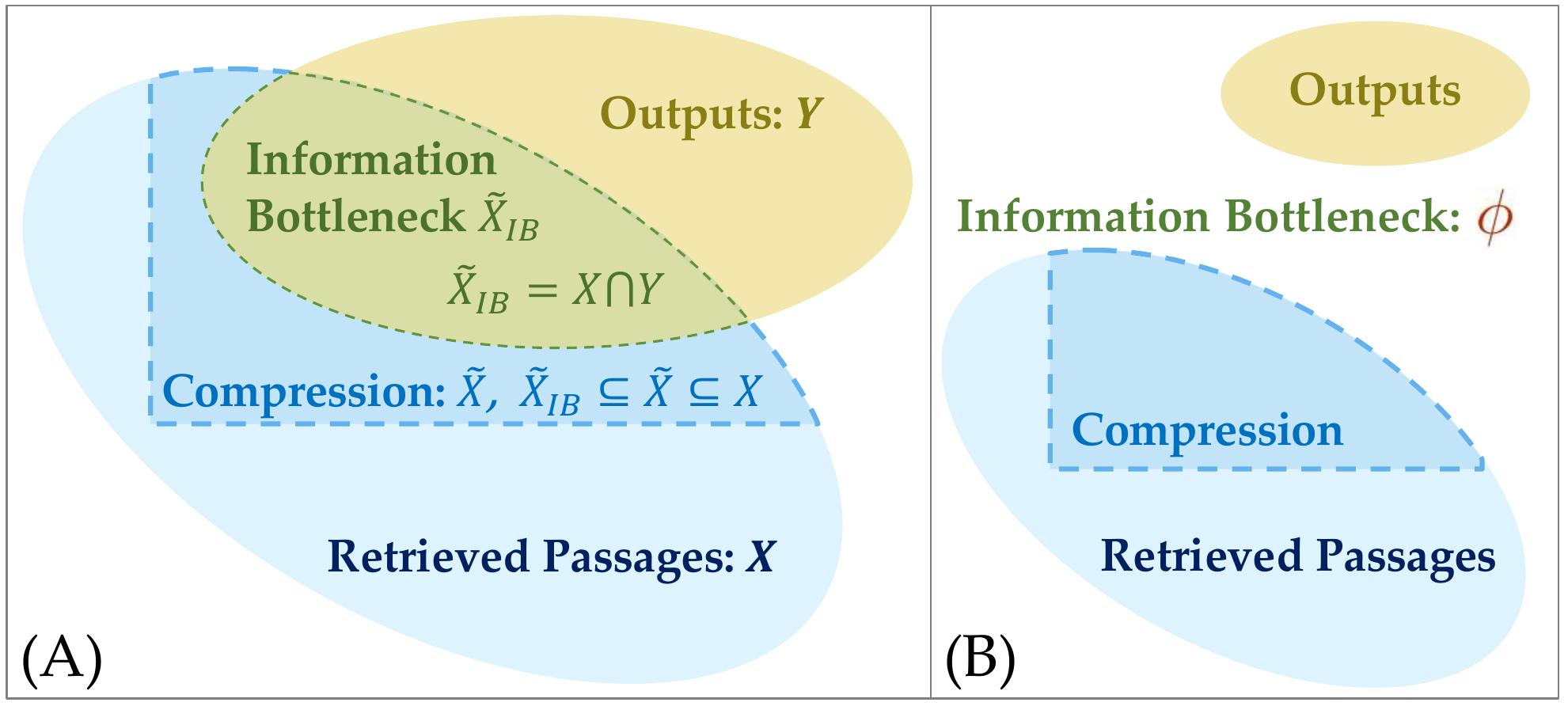}
    \caption{In retrieval-augmented generation, passages $X$ are retrieved to enhance the generation of output $Y$. (A) Recent Noise filtering approaches obtain the compression $\tilde{X}\subseteq X$ with log likelihood objective to outputs $Y$. Our information bottleneck objective enables a precise delineation of the intersection $\tilde{X}_{\exper{IB}}=X\cap Y$. (B) Information bottleneck explicitly compresses $\tilde{X}_{\exper{IB}}=\phi$, when retrieved passages are irrelevant to outputs.}
    \label{fig:information_bottleneck}
\end{figure}

Retrieval-augmented generation also presents problems. On the one hand, the retriever's efficacy may be suboptimal in practical use \cite{izacard2022few,shi2023replug,cheng2023uprise,lin2023ra}. On the other hand, the internet data is often of low quality, with redundancy and noise. Indeed, the retrieved content can be completely irrelevant to the query, leading to the model producing incorrect results \cite{shi2023large}. 
Recent solutions to mitigate noise in retrieval evidence often involve the adoption of a filtering module \cite{liu2023tcra, yang2023prca, xu2023recomp}. 
However, these methods encounter several issues:
(1) The inability to ensure that the annotated filtering results can effectively support the generation model in accurately answering questions.
(2) The difficulty in directing the filter to refrain from answering when confronted with retrieval evidence that does not support question resolution.
(3) The lack of adaptation to the compression extent of the filtering results, impeding the achievement of an optimal solution in terms of cost performance.







We observe that issues above originate from sub-optimal objectives.
As shown in Figure \ref{fig:information_bottleneck}A, the intersection between retrieved passages $X$ and outputs $Y$ denotes the precise information in $X$ which is useful for $Y$.
The noise filter extracts compression $\tilde{X}$ from retrieved passages $X$, where the filter is optimized with log likelihood objective to output $Y$.
The noise filter trained with this objective can obtain a compression $\tilde{X}$ containing the intersection $X\cap Y$, but is incapable of realizing its exact area, which means the filter cannot in principle eliminate the interference of noise for subsequent generation.
Therefore, we propose to utilize the information bottleneck theory \cite{tishby99information} to optimize the noise filter from a comprehensive perspective, via simultaneously maximizing the useful information while minimizing the noise, thus facilitating a precise delineation of the intersection $\tilde{X}_{\exper{IB}}=X\cap Y$. Furthermore, in cases (Figure \ref{fig:information_bottleneck}B) where retrieval is not necessitated for content generation or exhibits limited efficacy, the information bottleneck objective enables noise filters to compress the retrieved passages into empty $\tilde{X}_{\exper{IB}}=\phi$. 


Specifically, we consider information bottleneck as a principle for retrieval augmentation.
We first theoretically derive the formula of information bottleneck for retrieval-augmented generation, which integrates large language models.
Then we introduce information bottleneck as a new comprehensive evaluation metric for noise filtering, assessing both conciseness and correctness of compressed contents. 
Next we derive information bottleneck version of supervised fine-tuning and reinforcement learning objectives to train the noise filter.


We conduct experiments on the open-domain question answering datasets: Natural Questions (\exper{NQ}) \cite{kwiatkowski2019natural}, \exper{TriviaQA} \cite{joshi2017triviaqa}, and a more complex multi-hop \exper{HotpotQA} \cite{yang2018hotpotqa}.
Using \exper{Llama2} as filtering and generation model, our approach is proved to be effective compared with strong baseline models, including
\exper{RankGPT}, \exper{LongLLMLingua} and \exper{Lllama2} on all three datasets.
We achieve a $2.5\%$ significant compression rate and a $3.2$ improvement of exact answer match at most.



Our paper presents three main innovations:

\begin{itemize}
    \item We are the first, to the best of our knowledge, to introduce the information bottleneck theory into retrieval-augmented generation, which illustrates the optimum of filtration.
    \item We propose to apply the information bottleneck on evaluation metrics, supervised fine-tuning objectives, and reinforcement learning rewards for retrieval-augmented generation.
    \item Experimental results reveal the effectiveness of our approach in terms of generation correctness and compression conciseness.
\end{itemize}


\section{Related Work}

\paragraph{Information Bottleneck}

The Information Bottleneck (IB) \cite{tishby99information, fischer2020conditional} is a rather simplistic concept: when facing a task, one should attempt to accomplish it using minimal information. 
The Information Bottleneck theory characterizes learning as a delicate balance between data compression and information retention.
When applied to specific tasks, the idea is to extract all the essential informative features for the task while discarding redundant information \cite{shwartz2023compress}.
It has been applied in the study of representation learning \cite{wu2020graph, Federici2020Learning, lee2021compressive}, deep learning \cite{tishby2015deep, saxe2019information, kawaguchi2023does}, document clustering \cite{slonim2000document}, speech recognition \cite{hecht2009speaker}, summarization \cite{west2019bottlesum}, etc.

\begin{figure*}[t]
\centering
\includegraphics[width=2\columnwidth]{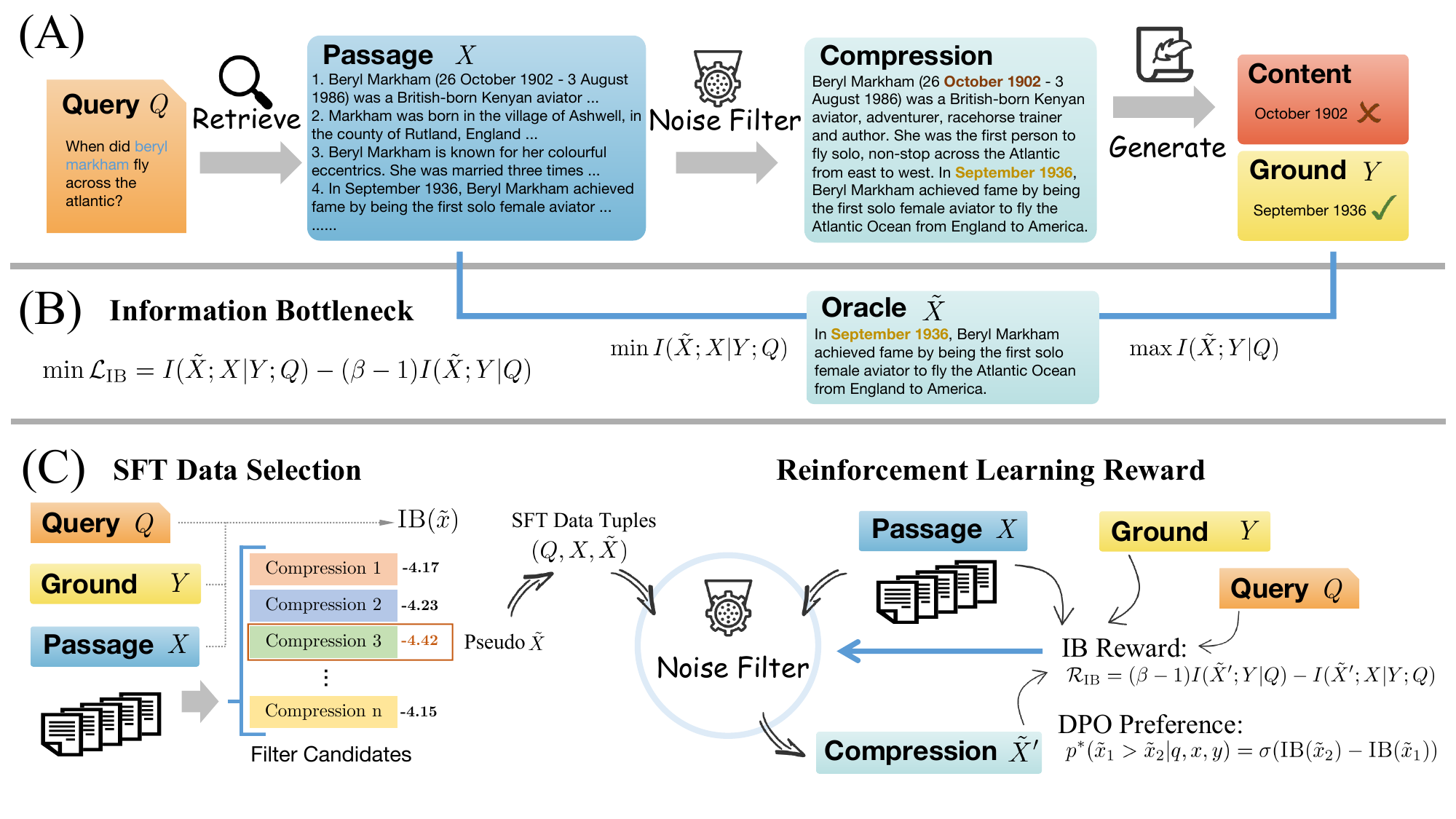}
\caption{Overview of our approach. (A) Traditional retrieval-augmented generation with noise filtering. Although the compression contains relevant information, the unavoidable noise continues to disrupt subsequent generation processes. (B) The information bottleneck theory provides an objective to oracle compression $\tilde{X}=X\cap Y$ via eliminating the influence of noise to the greatest extent, $\min I(\tilde{X};X|Y;Q)$. (C) We derive the formula for applying information bottleneck theory to retrieval-augmented generation, which can be utilized for the selection of supervised fine-tuning dataset and the construction of reinforcement learning reward.}
\label{fig:model}
\end{figure*}

\paragraph{Noise Filtering}

Retrieval-augmented generation usually concatenates retrieved passages with their queries as the input of language models.
However, this can potentially exceed the context window limit, introduce extra noise and redundancy, and increase computing resource requirements, leading to a decrease in model performance.

\exper{Flare} \cite{jiang2023active} and Self-\exper{Rag} \cite{asai2023self} are dedicated to training models to have the capability of actively retrieving and filtering retrieval content on their own.
\exper{Replug} \cite{shi2023replug} improves the retriever by the KL divergence between the retriever and the LLM.

Post-processing techniques such as noise filtering can help alleviate these issues.
\cite{bai2023griprank} focuses on re-ranking retrieved articles to filter out noise. 
Some methods like Selective Context \cite{li2023unlocking} and \exper{LLMLingua} \cite{jiang2023llmlingua} make use of small language models to measure prompt mutual information or perplexity, finding the highest-scoring elements. 

There are also some methods employ summarization techniques to design compressors \cite{xu2023recomp, wang2023learning}.
\exper{Tcra-LLM} \cite{liu2023tcra} and \exper{LongLLMLingua} \cite{jiang2023longllmlingua} combine summarization and semantic compression techniques.
\exper{Prca} \cite{yang-etal-2023-prca} also incorporates reinforcement learning algorithms in model training.
However, these compression methods do not have a unified evaluation of the compression results.
\exper{Recomp} \cite{xu2023recomp} achieves a compression rate of 6\%, but it comes at the cost of degraded performance.
Our approach is to find the optimal balance between compression rate and performance with the information bottleneck theory.

\section{Methodology} 

In this section, we first introduce the background of information bottleneck (\S \ref{sec:preliminary}), and then convert the information bottleneck into forms for noise filters of retrieval augmented generation (\S \ref{sec:noise_filter}). Next, we provide details of information bottleneck objectives for retrieval augmented generation (\S \ref{sec:objective}).
\subsection{Preliminary}
\label{sec:preliminary}
The information bottleneck principle \cite{tishby99information} has been a great concept in finding a compression $\tilde{X}$ for signals $X$ that preserves the maximum information relevant to signals $Y$. 
Given a joint probability distribution $p(X,Y)$ between a random variable $X$ and an observed relevant variable $Y$ (with support sets $x \in \mathcal{X}$ and $y \in \mathcal{Y}$), the amount of information about $Y$ in compressed representation $\tilde{X}$ ($\tilde{x} \in \tilde{\mathcal{X}}$) is given by mutual information:
\begin{equation}
    I(\tilde{X};Y)=\int_{\tilde{\mathcal{X}}}\int_{\mathcal{Y}} p(\tilde{x},y)\log\frac{p(\tilde{x}, y)}{p(\tilde{x})p(y)}\mathrm{d}\tilde{x}\mathrm{d}y,
\end{equation}
where $I(\tilde{X};Y)\leq I(X;Y)$ since compressed data can not convey more information than original ones.
The information bottleneck is obtained through
\begin{equation}
    \label{eq:IB}
    \min \mathcal{L}_{\text{IB}}=I(\tilde{X};X)-\beta I(\tilde{X};Y),
\end{equation}
where $\beta$ is the Lagrange multiplier for the trade-off between preserving meaningful information and compressing at various resolutions.

\subsection{Noise Filtering}
\label{sec:noise_filter}
Retrieval-augmented generation involves in generating contents conditioned on input queries $q\in Q$, where relevant passages $x\in X$ are retrieved to advance the content generation. 
As illustrated in Figure \ref{fig:model}A, recent noise filtering approaches for retrieval-augmented generation learn to compress retrieved passages $\tilde{x}\in \tilde{X} \subseteq X$ for large language models with a log likelihood objective $\minus\log p_{\text{LM}}(y|[q,\tilde{x}])$ to ground outputs $y\in Y$ \cite{liu2023tcra,yang2023prca,xu2023recomp,wang2023learning}, which are special cases of conditional mutual information $I(\tilde{X},Y|Q)$ and still unable to avoid irrelevant information.

Now we introduce our information bottleneck approach for retrieval-augmented generation and we derive the information bottleneck between retrieved passages $X$ and ground outputs $Y$ conditioned on queries $Q$ given Equation \ref{eq:IB}. As demonstrated in Figure \ref{fig:model}B, the noise filter is required to simultaneously maximizing the mutual information of its compression with ground outputs while minimizing the mutual information with retrieved passages:
\begin{equation}
    \min \mathcal{L}_{\text{IB}}=\underbrace{I(\tilde{X},X|Q)}_{\textit{conciseness}}-\beta \underbrace{I(\tilde{X}; Y|Q)}_{\textit{correctness}}.
\end{equation}
The former term $I(\tilde{X};X|Q)$ serves not only to enhance efficiency, which is a common application, but also to promote conciseness by minimizing the inclusion of irrelevant information. As the filtered information becomes increasingly precise, language models are able to reduce the computational resources allocated to input extraction, thereby enhancing their capacity to concentrate on producing higher-quality contents. It's worth noting that this term guarantees that the filtered information will be rendered null when the retrieved content is entirely irrelevant to the output.

Then we provide the details of each term in the information bottleneck. The \textit{conciseness} is:
\begin{equation}
    \begin{aligned}
        &I(\tilde{X};X|Q)=\\
        &\quad \, \, \mathbb{E}_q\bigg[\int p(\tilde{x},x|q)\log\frac{p(\tilde{x},x|q)}{p(\tilde{x}|q)p(x|q)}\mathrm{d}\tilde{x}\mathrm{d}x\bigg]\\
        &=\mathbb{E}_q\bigg[\int p(x|q)p(\tilde{x}|x,q)\\
        &\qquad\qquad \log\frac{p(x|\tilde{x},q) \bcancel{p(\tilde{x}|q)}}{p(x|q)\bcancel{p(\tilde{x}|q)}}\mathrm{d}\tilde{x}\mathrm{d}x\bigg]\\
        &\leq \mathbb{E}_q\bigg[\int p(x|q)\log \frac{\int p(x|\tilde{x},q)p(\tilde{x}|x,q)\mathrm{d}\tilde{x}}{p(x|q)}\mathrm{d}x\bigg]\\
        &= -\mathbb{E}_q\bigg[D_{\text{KL}}\Big[p(x|q)||\mathbb{E}_{\tilde{x}\sim p(\tilde{x}|x,q)} p(x|\tilde{x},q)\Big]\bigg],
    \end{aligned}
\end{equation}
where we get an upper bound of \textit{conciseness} based on Jensen's inequality. Therefore, $I(\tilde{X};X,Q)$ can be converted to the form of the Kullback–Leibler divergence between the retrieval probability distribution $p(x|q)$ and the expectation of the probability to recover retrieved passages from compression $p(x|\tilde{x},q)$, where $\tilde{x}$ is required to be integrated over the representation space of noise filter $p(\tilde{x}|x,q)$. 

In the scenario of offline retrievers, where the retrieved passages and queries are jointly sampled from training datasets $\{(q,x,y)\}$, $p(x|q)$ becomes a constant number and we can simplify:
\begin{equation}
    \label{eq:former_simplified}
    \min I(\tilde{X};X|Q)\simeq \min\mathbb{E}_{(q,x,\tilde{x})}\big[p(x|\tilde{x},q)\big].
\end{equation}
Hence, in cases where training the retriever jointly is not necessary, minimizing the conditional mutual information $I(\tilde{X};X|Q)$ can be elucidated as \textit{the process of selectively filtering out information to such an extent that it becomes unfeasible to reconstruct the original content, regardless of the strength of generative language models employed.}

Next, the \textit{correctness} is derived as:
\begin{equation}
    \begin{aligned}
        &I(\tilde{X};Y|Q)= H(Y|Q)-H(Y|X,Q)\\
        &=-\int p(y,q)\log p(y|q)\mathrm{d}y\mathrm{d}q -H(Y|X,Q)
    \end{aligned}
\end{equation}
where the former term $\mathbb{E}_{(q,y)}\big[\log p(y|q)\big]$ is considered as a constant independent of the noise filter. We can ignore this term and simplify as:
\begin{equation}
    \begin{aligned}
        &I(\tilde{X};Y|Q)\simeq-H(Y|\tilde{X},Q)\\
        &=\int p(y,\tilde{x},q)\log p(y|\tilde{x},q)\mathrm{d}y\mathrm{d}\tilde{x}\mathrm{d}q\\
        &=\mathbb{E}_{(q,\tilde{x},y)}\big[\log p(y|\tilde{x}, q)\big].
    \end{aligned}
\end{equation}
When generative language models are fixed, query-answer pairs $\{(q,y)\}$ are sampled from datasets and $\tilde{x}$ is pre-obtained with noise filter. Therefore, maximizing $I(\tilde{X};Y|Q)$ approximates maximizing log likelihood $\log p(y|\tilde{x},q)$, which is explained as \textit{the process of selectively retaining as much useful information as possible to enable the language model generate target outputs.}

Besides, recent studies on the information bottleneck \cite{fischer2020conditional, Federici2020Learning, lee2021compressive, pmlr-v202-kawaguchi23a} suggest to replace $I(X;\tilde{X})$ with $I(X;\tilde{X}|Y)$, because $I(X;\tilde{X})$ can not be zero while maintaining the target-relevant information.
Therefore, we follow \citet{Federici2020Learning} and decompose $I(\tilde{X};X|Q)$ into two components by using the chain rule as: 
\begin{equation}
    I(\tilde{X};X|Q) = I(\tilde{X};X|Y;Q) + I(\tilde{X};Y|Q).
\end{equation}

Finally, our information bottleneck for noise filtering in retrieval-augmented generation is:
\begin{equation}
    \label{eq:information_bottleneck_for_noise_filtering}
    \begin{aligned}
        &\mathcal{L}_{\text{IB}}=I(\tilde{X};X|Y;Q) - (\beta-1) I(\tilde{X};Y|Q)\\
        &\quad \simeq \mathbb{E}_{(q,x,\tilde{x},y)} \big[p(x|\tilde{x},q,y)\big]\\ 
        &\qquad - (\beta-1)\ \mathbb{E}_{(q,\tilde{x},y)}\big[\log p(y|\tilde{x},q)\big],
    \end{aligned}
\end{equation}
where the Lagrange multiplier $\beta-1 > 0$.
\subsection{Information Bottleneck as a Principle}
\label{sec:objective}
The information bottleneck represents not merely a methodological approach, but a fundamental principle to be applied in retrieval-augmented generation.
In this section, we will delineate three distinct applications of information bottlenecks, encompassing the establishment of an evaluation metric for noise filtering, the creation of supervised-fine-tuning (\exper{Sft}) training datasets, and the formulation of reward functions in reinforcement learning.
\subsubsection{Evaluation Metric}
Prior to describing the method for training a noise filter, it is imperative to establish criteria assessing the efficacy of filtration outcomes, where the information bottleneck serves as an important evaluation metric. Given $\{(q,x,y)\}$ from dataset and the compression generated by the noise filter $p(\tilde{x}|x,q)$, based on Equation \ref{eq:information_bottleneck_for_noise_filtering}, we define the \exper{IB} score as:
\begin{equation}
    \text{\exper{IB}}(\tilde{x}) = p_{\text{LM}}(x|[q,\tilde{x},y]) - \alpha\ p_{\text{LM}}(y|[q,\tilde{x}]),
\end{equation}
where large language models are employed to estimate probability distributions, with the inputs to the language model comprising concatenated conditional variables. In addition, we balance the magnitude of values by implying logarithms into the hyperparameter $\alpha$. The range of \exper{IB} score is $[\minus\alpha, 1]$ and smaller \exper{IB} means better filtration performance.

\subsubsection{Supervised Fine-tuning}
Training the noise filter from scratch is challenging since there is no ground truth compression of retrieved passages. Although Equation \ref{eq:information_bottleneck_for_noise_filtering} provides a way for achieving the oracle compression, we have to search the optimal one $\tilde{x}$ from all potential subsequences of the retrieved passage $x$, which is to calculate the integral of $\tilde{x}$ over the language space:
\begin{equation}
    \begin{aligned}
        \mathbb{E}_{(q,x,y)}\left[\int \exper{IB}(\tilde{x})\  p(\tilde{x}|x,q)\mathrm{d}\tilde{x}\right].
    \end{aligned}
\end{equation}
The integral is obviously intractable, but we can estimate it with the Monte Carlo sampling strategy. We utilize different existing compression or filtering approaches $\left\{p_{\theta_1}(\tilde{x}|x,q),\dots, p_{\theta_n}(\tilde{x}|x,q)\right\}$ to generate candidate compression outputs, as an approximation of sampling from $p(\tilde{x}|x,q)$, and the candidate of the best \exper{IB} score is considered as the pseudo $\tilde{x}$. As illustrated in the left part of Figure \ref{fig:model}C, we collect pseudo $\tilde{x}$ over the retrieval-augment generation datasets and construct the $\{(q,x,\tilde{x})\}$ tuples as training data for supervised learning of our noise filter $p_\theta(\tilde{x}|x,q)$. In addition, since the noise filter is required to possess capabilities of input understanding and instruction following, we choose pretrained language models as backbones and fine-tune the model to a silver noise filter. The optimization objective is commonly used negative log likelihood $\mathcal{L}_{\exper{Sft}}=-\sum_{(q,x,\tilde{x})}\log p_\theta(\tilde{x}|x,q)$.

It's worth noting that our approach shows a strong capability to handle the situation when retrieved passages $X$ are irrelevant to ground outputs $Y$ via minimizing $I(\tilde{X};X|Y;Q)\rightarrow 0$, which usually makes $\tilde{x}\rightarrow \phi$. 
Despite the fact that our information bottleneck objective inherently encompasses optimization objectives for addressing issues related to low-quality information, such as retrieval-free questions, noisy retrieval, and high-loss compression, we demonstrate the incorporation of an additional predictive flag \exper{[is\_discard]} to determine the necessity of discarding the current filtering outcomes. 
When $\exper{IB}(\phi) < \exper{IB}(\tilde{x})$, which means candidate compression contains too little useful information to assist in model generation, we will set $\exper{[is\_discard]}=\text{True}$ and vice versa.



\begin{table*}[t]
  \centering
\setlength\tabcolsep{3pt}
    \begin{tabular}{lcccccccccccc}
    \hline
    \multirow{2}[2]{*}{\textbf{Method}} & \multicolumn{5}{c}{\textbf{\exper{NQ}}} &   & \multicolumn{5}{c}{\textbf{\exper{TriviaQA}}} \\
       & \textbf{words} & \textbf{\exper{Em}}↑ & \textbf{\exper{Tfr}}↓ & \textbf{\exper{Ffr}}↑ & \textbf{\exper{F1}}↑ & \textbf{\exper{IB}}↓ & \textbf{words} &  \textbf{\exper{Em}}↑ & \textbf{\exper{Tfr}}↓ & \textbf{\exper{Ffr}}↑  & \textbf{\exper{F1}}↑ & \textbf{\exper{IB}}↓ \\
    \hline
    \multicolumn{5}{l}{\textbf{\textit{No Retrieval}}} \\
    \exper{Llama2-13B} & \phantom{00}0\phantom{.0} & 16.2  & - & - & 51.4  &  -4.46 &  \phantom{00}0\phantom{.0} & 49.9  & - & - & 76.7 & -4.68 \\
    \hline
    \multicolumn{5}{l}{\textbf{\textit{Retrieval without Noise Filtering}}} \\
    Top 1 document & 103.6 & 13.4 & 56.8 & \phantom{0}7.7 & 51.0  &  -4.29 & 102.9  & 46.5  & 28.3  & 21.4  & 75.7  &  -4.67\\
    Top 5 documents & 517.6 & 14.7  & 55.8  & \phantom{0}9.0  & 48.4  & -4.21  & 514.6 & 40.7  & 39.6  & 21.1  & 70.9  & -4.39\\
    \hline
    \multicolumn{5}{l}{\textbf{\textit{Retrieval with Filtering Method}}}\\
    \exper{RankGPT} & 103.6 & 16.5  & 51.0  & 10.3  & 53.7  & -4.47  & 102.8 & 47.5  & 28.3  & 23.3  & 76.1  & -4.70 \\
    \exper{LongLLMLingua} & 141.1 & 14.7  & 47.3  & \phantom{0}7.4  & 49.9  & -4.27  & 137.2 & 49.8  & 24.4  & 23.9  & 76.2  & -4.61 \\
    \exper{Llama2-7B} & \phantom{0}37.3 & 18.3  & 43.7  & \textbf{11.0}  & 52.2  & -4.53  & \phantom{0}30.0 & 51.4  & 23.1  & \textbf{26.0}  & 76.3  & -4.76 \\
    \hline
    Ours w/\exper{Sft} & \phantom{0}10.3 & 20.6 & \textbf{17.6}  & \phantom{0}8.7  &  54.9 & -4.60  & \phantom{0}11.6 & 50.3  & 14.6  & 15.2  & 77.4   & -4.79
    \\
    Ours w/\exper{Sft} w/\exper{Dpo} & \phantom{0}12.7 & \textbf{21.5} & 20.3 & 10.2 & \textbf{55.9} & \textbf{-4.78} & \phantom{0}13.3 & \textbf{52.1} & \textbf{12.5} & 16.8 & \textbf{78.2} &  \textbf{-4.88} \\
    \hline
    \end{tabular}%
    \caption{Open-domain QA results with \exper{Llama2-13B} as the generator. We report the \textbf{word} number of compressed retrieval evidence, which reflects the compression rate. Other evaluation metrics are in \S \ref{sec:metrics}.
    }\label{tab:main}%
\end{table*}%

\subsubsection{Reinforcement Learning}
Through merely supervised fine-tuning, the efficacy of the noise filtering is suboptimal, with its performance being constrained by the quality of compression candidates. Drawing parallels with the reinforcement learning from human feedback (\exper{Rlhf}, \citet{NEURIPS2022_b1efde53}) that aligns large language models with human preference, we propose leveraging reinforcement learning to enhance the noise filter by incorporating guidance from the information bottleneck. Our approach involves utilizing direct preference optimization (\exper{Dpo}, \citet{rafailov2023direct}), a recent iteration of \exper{Rlhf} that offers ease of implementation and robust stability. We define the preference probability as:
\begin{equation}
    p^*(\tilde{x}_1 > \tilde{x}_2|q,x,y) = \sigma(\exper{IB}(\tilde{x}_2) - \exper{IB}(\tilde{x}_1)),
\end{equation}
where our noise filter $p_\theta(\tilde{x}|x,q)$ is initially the policy $\pi_{\text{ref}}(\tilde{x}|x,q)$ that needs to generate two compression samples $\tilde{x}_1$ and $\tilde{x}_2$ before. Then the offline dataset of preferences $\mathcal{D} = \left\{q,x,\tilde{x}_w,\tilde{x}_l\right\}$ is constructed, where samples ($\tilde{x}_w$ wins and $\tilde{x}_l$ losses) are automatically labeled with the information bottleneck preferences $p^*(\tilde{x}_1 > \tilde{x}_2|q,x,y)$. Hence, the objective for the noise filtering policy $\pi_\theta(\tilde{x}|x,q)$, which is initialized with $\pi_{\text{ref}}(\tilde{x}|x,q)$, is as follows:
\begin{equation}
    \begin{aligned}
        &\mathcal{L}_{\exper{Dpo}} =\\
        &\ -\mathbb{E}_{(q,x,\tilde{x}_w,\tilde{x}_l)\sim \mathcal{D}} \bigg[\log \sigma\bigg(\gamma\log\frac{\pi_\theta(\tilde{x}_w|x,q)}{\pi_{\text{ref}}(\tilde{x}_w|x,q)}\\
        &\qquad -\gamma\log\frac{\pi_\theta(\tilde{x}_l|x,q)}{\pi_{\text{ref}}(\tilde{x}_l|x,q)}\bigg)\bigg],
    \end{aligned}
\end{equation}
where $\gamma$ is a hyperparameter controlling the deviation from the base reference policy $\pi_{\text{ref}}(\tilde{x}|x,q)$.
As shown in the right part of Figure \ref{fig:model}C, our information bottleneck can also provide the reward function $\mathcal{R}_{\exper{IB}}(\tilde{x})=-\exper{IB}(\tilde{x})$ of online policies.

\section{Experiments}

\subsection{Experimental Settings}

\paragraph{Datasets and Retrieval Corpus} 
We conduct experiments on three question answering benchmarks: Natural Questions (\exper{NQ}) \cite{kwiatkowski2019natural}, \exper{TriviaQA} \cite{joshi2017triviaqa} and \exper{HotpotQA} \cite{yang2018hotpotqa}, with results reported on development sets. 
We utilize the adversarial Dense Passage Retriever\footnote{\href{https://rgw.cs.uwaterloo.ca/pyserini/indexes/lucene-index.wikipedia-dpr-100w.20210120.d1b9e6.tar.gz}{Lucene index of Wikipedia with DPR 100-word splits}} (\exper{Dpr}) \cite{karpukhin-etal-2020-dense} to retrieve the top 5 passages from all Wikipedia passages for all datasets. 
The articles are truncated into non-overlapping documents of 100 words.  

\begin{table}[htbp]
\setlength\tabcolsep{4.8pt}

\begin{tabular}{lcccc}
\cline{1-5}\multirow{2}[2]{*}{Dataset} & \multirow{2}[2]{*}{Type} & \multirow{2}[2]{*}{Examples} & \multicolumn{2}{c}{\% of Recall } \\
  &   &   & Top 5 & Top 1 \\
\cline{1-5}\multirow{2}[2]{*}{\exper{NQ}} & Train & 106926 & 36.5 & 18.9 \\
  & Dev & \phantom{00}2564 & 35.0 & 17.9 \\
\cline{1-5}\multirow{2}[2]{*}{\exper{TriviaQA}} & Train & \phantom{0}87622 & 68.3 & 49.1 \\
  & Dev & \phantom{0}11313 & 68.5 & 49.4 \\
\cline{1-5}\multirow{2}[2]{*}{\exper{HotpotQA}} & Train & \phantom{0}90447 & 51.5 & 35.1 \\
  & Dev & \phantom{00}7405 & 45.0 & 28.4 \\
\cline{1-5}\end{tabular}%
\caption{Overview of the data quantities used for training and testing across three benchmark datasets and recall of the top 5 and top 1 \exper{Dpr}-retrieved passages.}
\end{table}%

\paragraph{Implementation Details} 
We use \exper{Llama2} \cite{touvron2023llama} as the backbone architecture of the large language model. 
We finetune the 7B model version with \exper{LoRA} \cite{hu2021lora} for noise filtering, where the optimizer is \exper{AdamW} with the learning rate of 5e-5 and the batch size of 32.
The 13B version is also employed as the generator without any adjustments. For supervised fine-tuning, we set the coefficient of \exper{IB} score $\alpha=10$ to balance the compression. 

Besides, filtering candidates are sampled from four different methods derived from traditional extractive summarization, with details in \S \ref{sec:dev-oracle}. For reinforcement learning, we utilize \exper{Dpo} and the hyperparameter $\gamma=0.1$.
The decoding strategy is top-p sampling with $p=0.9$.


\paragraph{Metrics}
\label{sec:metrics}
Question-answering tasks typically employ Exact Match (\textbf{\exper{Em}}) and \textbf{\exper{F1}} as evaluation metrics, while each of them has its own limitations.
For example, the \exper{Em} metric requires an exact alignment between the generated content and the answer, exhibiting an over-strict criterion that lacks semantic compatibility.
Besides, the \exper{F1} score, functioning as an uni-gram metric, is easy to be cheated by negative terms such as ``no.''
Our \textbf{\exper{IB}} score, on the contrary, is a comprehensive and versatile evaluation metric capable of assessing the \textit{conciseness} and \textit{correctness} of generated contents on a semantic level compared to ground answers, leveraging the capabilities of advanced large language models.
In addition, to evaluate the performance variations resulting from retrieval augmentation in detail, we use the flip rate of \exper{Em} to evaluate the extent to which generated responses are influenced by retrieved context. The True-Flip-Rate (\textbf{\exper{Tfr}}) and False-Flip-Rate (\textbf{\exper{Ffr}}) are defined as follows:
\begin{equation}
    \nonumber
    \begin{aligned}
    \exper{Tfr}&=p(EM_{p_{\text{LM}}(y|[q,x])} = 0 | EM_{p_{\text{LM}}(y|q)} = 1)\\
    \exper{Ffr}&=p(EM_{p_{\text{LM}}(y|[q,x])} = 1 | EM_{p_{\text{LM}}(y|q)} = 0),
    \end{aligned}
\end{equation}
where \exper{Tfr} measures the degree of noise introduced by the retrieved information while \exper{Ffr} examines the amount of benefits from retrieval augmentation.

\paragraph{Baselines} 

First we consider the results of the generator (\exper{Llama2-13B}) without retrieval augmentation and with top-1 or top-5 retrieved passages.
We next include two sets of filtering methods, reranking using \exper{RankGPT} \cite{sun2023chatgpt} distilled from ChatGPT, and prompt compression with \exper{LongLLMLingua} \cite{jiang2023longllmlingua}.
We also experiment with the base model \exper{Llama}-7B doing summarization from top-5 passages.





\subsection{Open-Domain Question Answering}

Table \ref{tab:main} presents the experimental results on \exper{NQ} and \exper{TriviaQA} datasets.
Initially, we demonstrate the performance of the generator (\exper{Llama2 13B}) without retrieval, where a low \exper{Em} score implies the limitation of large language models facing open-domain question answering task. When the generation is directly augmented with retrieved documents, the result is even worse with a drop of $9.2$ \exper{Em} score on \exper{TriviaQA} at most, reflecting the presence of a large amount of noise in the retrieved content and the vulnerability of the generator to noise interference. Besides, the \exper{Tfr} reveals that noise in retrieved passages will cause over $50\%$ of the generated answers wrong that could have been answered correctly on the \exper{NQ} dataset.

Extractive noise filters including \exper{RankGPT} and \exper{LongLLMLingua} can compensate for the performance losses in \exper{Em}, \exper{F1}, and \exper{IB} scores caused by retrieval, via noise filtering. 
However, they can hardly outperform the non-retrieval generator, because they enable the generator to answer more questions correctly with $10.3$ of \exper{Ffr}, but also make more mistakes with $51.0$ of \exper{Tfr}. Also as an extractive compression method, our approach significantly improves the \exper{Em} scores on on \exper{NQ} and \exper{TriviaQA} by $5.3$ and $2.2$, compared with the non-retrieval generator. For \exper{F1} scores, the advancement are $4.5$ and $1.5$ respectively. Our information bottleneck objectives can achieve considerable \exper{Ffr} improvement while minimizing \exper{Tfr} performance degradation. 
For instance, on \exper{NQ} dataset, we effectively minimize \exper{Tfr} ($51.0\rightarrow 17.6$) to alleviate the noise interference caused by retrieval, and maintain a comparable \exper{Ffr} ($10.3\rightarrow 10.2$) to minimize the loss of effective information as much as possible.
It's worth noting that our compression rate achieves $2.5\%$ and $2.6\%$ for these datasets, 
which impressively reduces irrelevant noise and computational cost. 
Compared to \exper{Llama2}-7B, a large language model with knowledge stored in parameters during pretraining stage where we consider it as a powerful abstractive summarization method, our method still outperforms it in \exper{Em} and \exper{F1} scores.

In addition, our proposed information bottleneck metric \exper{IB} assesses these models from a comprehensive perspective, which takes both conciseness and correctness of compression into account. The \exper{IB} score of non-retrieval is the basic score for null compression $\tilde{X}=\phi$, which reflects the capability of the language model $p_{\text{LM}}(\cdot)$. The \exper{IB} score generally aligns with the model's performance, where models exhibiting lower compression rates and reduced accuracy tend to yield lower scores.
Our approach, with the best \exper{IB} score, achieves the Pareto optimum between compression ratio and performance. We provide case study in Appendix \ref{sec:case-study}.

\begin{table}[t]
  \centering
  \setlength\tabcolsep{2pt}
     \begin{tabular}{l|cccccc}
    \hline
    \multirow{2}[2]{*}{\textbf{Method}} & \multicolumn{5}{c}{\textbf{\exper{HotpotQA}}} \\
      & \textbf{words} & \textbf{\exper{Em}} & \textbf{\exper{Tfr}} & \textbf{\exper{Ffr}} & \textbf{\exper{F1}} & \textbf{\exper{IB}} \\
    \hline
    \small{\exper{Llama2}-13B} & \phantom{00}0\phantom{.0} & 18.5  & - & - & 53.6 & -4.21 \\
    \small{Top 1 document} & 102.9 & 23.2 & 31.1  & 12.8  & 57.3  & -4.36  \\
    \small{Top 5 documents} & 514.7 & 18.3  & 50.7  & 11.2  & 50.2  &  -4.10\\
    \hline
    \small{\exper{RankGPT}} & 102.9 & 23.5  & 34.1  & 13.9  & 57.2 & -4.39 \\
    \scriptsize{\exper{LongLLMLingua}} & 137.7 & 23.9  & 32.7  & 14.0  & 56.4  & -4.19  \\
    \exper{Llama2} & \phantom{0}27.7 & 25.9  & 31.1  & \textbf{16.2}  & 57.8  & -4.44  \\
    \hline
    Ours  & \phantom{0}13.2 & \textbf{26.1} & \textbf{22.1} & 14.3 & \textbf{58.3}  & \textbf{-4.47} \\
    \hline
    \end{tabular}%
    \caption{Results on the multi-hop \exper{HotpotQA} dataset. }
  \label{tab:multihop}%
\end{table}%



\begin{table}[t]
  \centering
  \setlength\tabcolsep{2.8pt}
    \begin{tabular}{lcccccc}
    \hline
      & \textbf{words} & \textbf{\exper{Em}}↑ & \textbf{\exper{Tfr}}↓ & \textbf{\exper{Ffr}}↑ & \textbf{\exper{F1}}↑ & \textbf{\exper{IB}}↓ \\
    \hline
    \small{\exper{Llama2}} & - & 16.2  & - & - & 51.4  & -4.46 \\
    Top1 & 103.6  & 13.4  & 56.8  & \phantom{0}7.7  & 51.0  & -4.29 \\
    Top5 & 517.6  & 14.7  & 55.8  & \phantom{0}9.0  & 48.4  & -4.21 \\
    \scriptsize{$I(\tilde{X};Y|Q)$} & \phantom{0}13.1  & 19.2  & 24.9  & \phantom{0}8.5  & 54.6  & -4.58 \\
    \exper{IB} & \phantom{0}\textbf{12.7}  & \textbf{21.5}  & \textbf{20.3}  & \textbf{10.2}  & \textbf{55.9}  & \textbf{-4.78} \\
    \hline
    \end{tabular}%
    \caption{Ablation study for \textit{conciseness} on \exper{NQ}.}
  \label{tab:ablation}%
\end{table}%

\begin{table*}[ht]
  \centering
  \setlength\tabcolsep{4.5pt}

    \begin{tabular}{c|ll|cccccc}
    \hline
    \multicolumn{1}{c}{\textbf{Dataset}} &   \multicolumn{2}{c}{\textbf{Filtering Candidates}}  & \textbf{\exper{HasAns}} & \textbf{\exper{Em}} & \textbf{\exper{F1}} & \textbf{Words} & \textbf{\exper{IB}} & \footnotesize{$I(\tilde{X};X|Y;Q)$}\\
    \hline
    \multirow{5}[2]{*}{NQ} & \multicolumn{1}{l}{\multirow{2}[1]{*}{Exact}} 
          & Paragraph-Level & 31.4 & 21.2 & 53.6 & 78.1 & -4.74 & 0.597\\
      &   & Senetence-Level & 33.5 & 23.8 & 55.4 & 28.4 & -4.81 & 0.561\\
      & \multicolumn{1}{l}{\multirow{2}[0]{*}{Greedy}} 
          & Query \& Answer & 26.6 & 19 & 52.1 & 26.2 & -4.64 & 0.562\\
      &   & Answer & 34.2 & 24.3 & 56.8 & 18.2 & -4.91 & 0.556\\
      & \multicolumn{2}{l|}{\exper{IB} Selection} & \textbf{35.7} & \textbf{26.8} & \textbf{58.6} & 31.6 & -5.10 & 0.563 \\
    \hline
    \multirow{5}[2]{*}{HotpotQA} & \multicolumn{1}{l}{\multirow{2}[1]{*}{Exact}} 
          & Paragraph-Level & 38.3 & 26.3 & 55.4 & 120.0 & -4.55 & 0.679 \\
      &   & Senetence-Level & 35.4 & 27.8 & 59.3 & 41.2 & -4.63 & 0.619 \\
      & \multicolumn{1}{l}{\multirow{2}[0]{*}{Greedy}} 
          & \scriptsize{Query \& Supporting Facts \& Answer} & 31.4 & 25.8 & 58.9 & 32.5 & -4.51 & 0.614 \\
      &   & \scriptsize{Supporting Facts \& Answer} & 33.1 & 26.9 & 59.5 & 14.8 & -4.63 & 0.604\\
      & \multicolumn{2}{l|}{\exper{IB} Selection} & \textbf{38.3} & \textbf{30.9} & \textbf{61.9} & 40.4 & -4.88 & 0.619  \\
    \hline
    \end{tabular}%
      \caption{We validate the effectiveness of the information bottleneck in finding the oracle filtered data on the dev sets of NQ and TriviaQA. \exper{HasAns} denotes the accuracy of filtered results having the answer.}
  \label{tab:dev-oracle}%
\end{table*}%

\subsection{Multi-Hop Question Answering}

Filtering models encounter more significant hurdles when dealing with multi-hop problems, as solving these requires not only auxiliary information but multiple rounds of reasoning and analysis.
As demonstrated in Table \ref{tab:multihop}, the results obtained are marginally improved compared to those achieved through supervised training alone.


\subsection{Ablation Study for \textit{Conciseness}}

We utilize ablation experiments to exhibit the significance of the \textit{conciseness} term in the information bottleneck theory on \exper{NQ}. As shown in Table \ref{tab:ablation}, integrating the information bottleneck approach that combines both \textit{conciseness} and \textit{correctness} leads to superior outcomes compared to solely utilizing \textit{correctness}.
The filtered outcomes from the former approach are more concise and of superior quality.

\section{Analysis}



\subsection{Silver Selection with \exper{IB} }
\label{sec:dev-oracle}
To visually demonstrate that applying the information bottleneck to select training data can enhance the upper bound, we conducted experiments on the validation sets of \exper{NQ} and \exper{HotpotQA}.
Table \ref{tab:dev-oracle} lists two basic filtering methods: exact search and greedy search. 
The application of \exper{IB} selection is not limited to just two filtering methods and it can be generalized to other answer-guided silver mining strategies, such as the leave-one-out evidentiality mining strategy \cite{asai2022evidentiality} and \exper{Cxmi} \cite{wang2023learning}. 
Our approach is not fixated on finding the optimal solution in the initial stages, but rather focuses on gradually approaching the optimal filter model through iterative training.
Here we choose the two easiest methods, which can greatly reduce computational costs on the construction of training data

The goal of exact search is to find the paragraphs or sentences containing the ground answers. 
Greedy search \cite{nallapati2017summarunner} is one of the most popular heuristic method by far used in extractive summarization. This algorithm extracts oracle labels with the highest \exper{Rouge} \cite{lin-2004-rouge} scores compared to human-annotated abstracts. We considered two silver summaries, one that concatenates the query and answer, and the other that focuses solely on the answer itself. 
The former can cover more information, while the latter focuses more on the answer itself.
Specially, the answer in intermediate state, supporting facts, are incorporated for multi-hop questions.



By using information bottleneck for selecting among the four filtering results, the obtained outcome is superior to any of them, which reveals that existing simple annotation methods only yield solutions that are far from optimal.

\subsection{Length of Summary and \textit{Conciseness}}

Due to the adoption of extractive filtering method, the content retained after filtering is sourced from the original text. 
This leads to the hypothesis that there may be a correlation between the compression rate and the \textit{conciseness} mutual information $I(\tilde{X};X|Y;Q)$.
We verify the relationship between compression rate and mutual information on \exper{NQ} and \exper{HotpotQA} with a toy experiment based on \S \ref{sec:dev-oracle}.
For each query-answer pair with four filtering candidates, we calculate their corresponding length and \textit{conciseness} $I(\tilde{X};X|Y;Q)$.
As the statistical information for different samples is independently and identically distributed, we convert the values of length and \textit{conciseness} in each sample into their rankings among various compression methods.
Then we calculate the Pearson correlation coefficient between length $Rank_{L}$ and \textit{conciseness} $Rank_{C}$, which is $0.953$, indicating a significant correlation at the 0.01 level (two-tailed).



\section{Conclusion}


We apply the information bottleneck principle to noise filters in retrieval-augmented generation, balancing the trade-off between the conciseness and correctness.
Not only as an evaluation method for noise filtering, we also apply it to select supervised fine-tuning training data and provide reward for reinforcement learning.
Codes are available at \href{https://github.com/zhukun1020/NoiseFilter\_IB}{https://github.com/zhukun1020/NoiseFilter\_IB}.
Our two-stage optimization gradually approaches the oracle filtering objective.
Experimental results demonstrate that our filter significantly outperforms baseline methods and achieves an impressive compression rate.

\section*{Acknowledgements}
Kun Zhu and Xiaocheng Feng contribute equally to to this work.
Bing Qin is the corresponding author of this work. We thank the anonymous reviewers for their insightful comments. This work was supported by the National Key R\&D Program of China via grant No. 2021ZD0112905, National Natural Science Foundation of China (NSFC) via grant 62276078 and U22B2059, the Key R\&D Program of Heilongjiang via grant 2022ZX01A32, the International Cooperation Project of PCL, PCL2022D01 and the Fundamental Research Funds for the Central Universities (Grant No.HIT.OCEF.2023018).

\section*{Limitations}

Although our method has exhibited effectiveness in enhancing the performance of noise filtering task on retrieval-augmented generation, it does have limitations such as performance reliance on the generator and trade-off between True-Flip-Rate (\exper{Tfr}) and False-Flip-Rate (\exper{Ffr}).
In order to analyze the mutual information between filtered content $\tilde{X}$ and retrieval content $X$, as well as between $\tilde{X}$ and $Y$, it is crucial to utilize a white-box generator equipped with robust capabilities.
Furthermore, by introducing the additional predictive flag to assess the necessity of discarding the current filtering outcomes, we successfully decrease \exper{Tfr}, while at a cost of potentially decreasing \exper{Ffr}.
We mitigate it by engaging in training iterations, which inevitably leads to an escalation in training cost.

\section*{Ethics Statement}
We are totally aware that text generation technology has a potential to be used maliciously to generate fake, toxic, or offensive content. 
If the retrieved content includes harmful or toxic information, it will influence the output of the generated content.
Our approach is proposed to mitigate the influence of noise from retrieval, which includes toxic contents. However, there is no assurance that our approach will completely eliminate toxics.

\bibliography{custom}

\appendix

\section{Prompt}
\subsection{Filter Prompt}

We show prompts used to train the filter in Table \ref{tab:noise}.

\begin{table}[ht]
    \centering
    \begin{tabular}{|p{7.5cm}|}
        \hline  
        
        \hline
        \text{[INST]} \\
        <\!<SYS>\!> \\
        You are now an intelligent assessment assistant. Your task is to read the context and then find coherent excerpts that can effectively answer the given question.After generating the answer, you need to determine whether the generated excerpt contributes to addressing the question. \\
<\!</SYS>\!> \\
Question: \{\} \\
Context: \\ 
\{\} \\
\text{[/INST]} \\
Question: \{\} \\
Excerpt: \{\} \\
Contribution: [\{\}]\\       

        \hline
        
        \hline
    \end{tabular}
    \caption{Filter Prompt}
    \label{tab:noise}
\end{table}

\subsection{Generator Prompt}

We show prompts used for inference in Table \ref{tab:gen}.
\begin{table}[ht]
    \centering
    \begin{tabular}{|p{7.5cm}|}
        \hline  
        
        \hline
        \text{[INST]} \\
        <\!<SYS>\!> \\
You are a helpful, respectful and honest assistant. 
Your task is to predict the answer to the question based on the given context. 
If you don't know the answer to a question, please don't share false information.
Answer the question as accurately as possible and put the answer in the form [answer]. \\
\\
Here is an example: \\
    Question: Who was the first person killed in a car accident? \\
    Answer: [Bridget Driscoll] \\
\\
    Question: Are both The New Pornographers and Kings of Leon American rock bands?  \\ 
    Answer: [no] \\
\\
    Question: What is the length of the track where the 2013 Liqui Moly Bathurst 12 Hour was staged? \\
    Answer: [6.213 km long] \\
\\
    Question: Which was the first European country to abolish capital punishment?    \\
    Answer: [Norway] \\
\\
(END OF EXAMPLE) \\
<\!</SYS>\!> \\
Given the ['question', 'context'], predict the answer to the question. \\

Question: \{\} \\
Context: \\
\{\} \\

[/INST] \\
Answer: [\{\}]\\        
\hline

\hline
    \end{tabular}
    
    \caption{Generator Prompt}
    \label{tab:gen}
    
\end{table}

\section{Experimental Configuration}
We use \exper{Llama2} \cite{touvron2023llama} as the backbone architecture of the large language model. 
We fine-tune the 7B model version with \exper{LoRA} \cite{hu2021lora} for noise filtering, where the optimizer is \exper{AdamW} with the learning rate of 5e-5 and the batch size of 32.
The 13B version is also employed as the generator without any adjustments. 

For supervised fine-tuning, we set the coefficient of \exper{IB} score $\alpha=10$ to balance the compression. 
As the value of $\alpha$ decreases, there is a greater tendency to prioritize concise filtering results as the silver selection. Conversely, as the value of increases, there is a greater inclination to prioritize higher correctness. We balance the conciseness and correctness by the hyperparameter.
In section \ref{sec:dev-oracle}, we conduct experiments on the validation sets of NQ and HotpotQA to visually demonstrate that applying the information bottleneck to select training data can enhance the upper bound. We also conduct experiments about the value of $\alpha$ here. As shown in the table \ref{tab:alpha} below, as long as the value of $\alpha$ is not zero, its value does not significantly impact the other performance metrics , such as HasAnswer, EM, and F1. $\alpha=10$ is a good empirical value that balance the conciseness and correctness. 

Besides, filtering candidates are sampled from four different methods derived from traditional extractive summarization, with details in \S \ref{sec:dev-oracle}. For reinforcement learning, we utilize \exper{Dpo} and the hyperparameter $\gamma=0.1$.
We use one NVIDIA-A100-40G to train 5 epochs given the question and 5 retrieval passages.
For every additional 1W of data, the training time increases by 5.5 hours.
We also use one NVIDIA-A100-80G to generate the answer given the question and compression results.
We set a maximum length of 1024 tokens for all sequences during training and inference. 
The generator is configured to generate a maximum of 200 tokens and the decoding strategy is top-p sampling with $p=0.9$.

\begin{table*}[htbp]
  \centering
    \begin{tabular}{r|rrrrrrrr}
    \hline
    \textbf{Dataset} & \textbf{$\alpha$} & \textbf{HasAns$\uparrow$} & \textbf{EM$\uparrow$} & \textbf{F1$\uparrow$} & \textbf{Words} & \textbf{IB$\downarrow$} & \textbf{loss1$\downarrow$} & \textbf{loss2$\uparrow$} \\
    \hline
    NQ & 0 & 29.21 & 21.10 & 54.19 & 7.01 & 0.548 & 0.5481 & 0.5244 \\
      & 1 & 35.69 & 26.79 & 58.67 & 16.46 & -0.009 & 0.5541 & 0.5631 \\
      & 2 & 36.08 & 27.11 & 58.74 & 19.74 & -0.573 & 0.5561 & 0.5645 \\
      & 3 & 36.19 & 27.11 & 58.81 & 22.05 & -1.138 & 0.5579 & 0.5652 \\
      & 4 & 36.00 & 26.95 & 58.63 & 24.97 & -1.703 & 0.5595 & 0.5657 \\
      & 5 & 35.88 & 26.99 & 58.68 & 26.43 & -2.269 & 0.5604 & 0.5659 \\
      & 6 & 35.8 & 26.95 & 58.66 & 27.53 & -2.835 & 0.5612 & 0.5661 \\
      & 7 & 35.69 & 26.87 & 58.62 & 28.68 & -3.401 & 0.5618 & 0.5661 \\
      & 8 & 35.61 & 26.79 & 58.53 & 29.70 & -3.967 & 0.5623 & 0.5662 \\
      & 9 & 35.73 & 26.83 & 58.59 & 30.38 & -4.534 & 0.5627 & 0.5663 \\
      & \textbf{10} & \textbf{35.73} & \textbf{26.79} & \textbf{58.56} & \textbf{31.63} & \textbf{-5.100} & \textbf{0.5635} & \textbf{0.5663} \\
      & 11 & 35.73 & 26.79 & 58.53 & 31.87 & -5.667 & 0.5637 & 0.5664 \\
      & 12 & 35.65 & 26.87 & 58.57 & 32.73 & -6.233 & 0.5642 & 0.5663 \\
      & 13 & 35.65 & 26.87 & 58.57 & 33.95 & -6.799 & 0.5649 & 0.5665 \\
      & 14 & 35.73 & 26.95 & 58.63 & 34.31 & -7.365 & 0.5652 & 0.5664 \\
      & 15 & 35.69 & 26.91 & 58.61 & 34.49 & -7.932 & 0.5653 & 0.5665 \\
      & 16 & 35.73 & 26.95 & 58.63 & 34.93 & -8.498 & 0.5657 & 0.5665 \\
      & 17 & 35.76 & 26.95 & 58.64 & 35.23 & -9.065 & 0.5659 & 0.5665 \\
      & 18 & 35.76 & 26.95 & 58.64 & 35.33 & -9.632 & 0.5660 & 0.5665 \\
      & 19 & 35.80 & 26.95 & 58.63 & 35.90 & -10.198 & 0.5664 & 0.5665 \\
      & 20 & 35.80 & 26.95 & 58.61 & 35.94 & -10.765 & 0.5664 & 0.5665 \\
    \hline
    HotpotQA & 0 & 31.74 & 25.39 & 57.77 & 9.46 & 0.600 & 0.6000 & 0.5132 \\
      & 1 & 37.25 & 30.21 & 61.52 & 19.86 & 0.061 & 0.6058 & 0.5451 \\
      & 2 & 37.61 & 30.49 & 61.70 & 24.91 & -0.486 & 0.6089 & 0.5472 \\
      & 3 & 37.84 & 30.70 & 61.91 & 28.46 & -1.033 & 0.6112 & 0.5482 \\
      & 4 & 37.97 & 30.80 & 61.96 & 31.30 & -1.582 & 0.6130 & 0.5487 \\
      & 5 & 38.11 & 30.93 & 62.00 & 33.37 & -2.131 & 0.6144 & 0.549 \\
      & 6 & 38.15 & 30.90 & 62.00 & 34.97 & -2.680 & 0.6154 & 0.5492 \\
      & 7 & 38.20 & 30.90 & 61.98 & 36.60 & -3.229 & 0.6165 & 0.5494 \\
      & 8 & 38.22 & 30.89 & 61.93 & 37.94 & -3.779 & 0.6173 & 0.5485 \\
      & 9 & 38.27 & 30.91 & 61.89 & 39.30 & -4.328 & 0.6183 & 0.5496 \\
      & \textbf{10} & \textbf{38.31} & \textbf{30.91} & \textbf{61.90} & \textbf{40.43} & \textbf{-4.878} & \textbf{0.6190} & \textbf{0.5497} \\
      & 11 & 38.33 & 30.86 & 61.89 & 41.42 & -5.428 & 0.6197 & 0.5498 \\
      & 12 & 38.38 & 30.87 & 61.91 & 42.59 & -5.977 & 0.6205 & 0.5498 \\
      & 13 & 38.35 & 30.82 & 61.83 & 43.64 & -6.527 & 0.6213 & 0.5499 \\
      & 14 & 38.38 & 30.80 & 61.82 & 44.54 & -7.077 & 0.6220 & 0.5499 \\
      & 15 & 38.39 & 30.75 & 61.80 & 45.22 & -7.627 & 0.6225 & 0.5500 \\
      & 16 & 38.43 & 30.75 & 61.79 & 46.23 & -8.177 & 0.6231 & 0.5500 \\
      & 17 & 38.42 & 30.68 & 61.73 & 46.89 & -8.727 & 0.6237 & 0.5500 \\
      & 18 & 38.43 & 30.70 & 61.74 & 47.52 & -9.277 & 0.6241 & 0.5500 \\
      & 19 & 38.45 & 30.70 & 61.72 & 48.33 & -9.827 & 0.6247 & 0.5501 \\
      & 20 & 38.46 & 30.70 & 61.72 & 49.04 & -10.377 & 0.6252 & 0.5501 \\
    \hline
    \end{tabular}%
    \caption{Experiments about the value of $\alpha$.}
  \label{tab:alpha}%
\end{table*}%

\section{Case Study}
\label{sec:case-study}

In this section, we present exxamples from each of the three datasets within two distinct scenarios. 
The first scenario is when the retrieved content fails to directly address the question and the compressed content should be empty.
Alternatively, the second scenario involves instances where the retrieved content effectively addresses the research question, and the compressed content should be as short as possible.

\begin{table*}
    \centering
    \begin{tabular}{|p{15.6cm}|}
        \hline  
        
        \hline
        \textbf{Question}: who won the 2017 sports personality of the year \\
        \textbf{Standard Answer}: Mo Farah \\
        \hline \hline
        \textbf{Retrienval Passages} \\
        "Newsfirst Platinum Awards"\newline Angelo Mathews won the Most Popular Sports Person of The Year Award. The second Platinum ceremony launching started with January 2017, with the full support of the Ministry of Education. Nominations for Platinum Awards 2017 opened under 20 categories. The promotional campaign commenced at Thurstan College, Colombo on 30 January 2017. The campaign was held in every district in the island until the 28 February 2017. The Awards night was held on 31 March 2017 at Sirasa Stein Studios, Ratmalana. Five times world champion in Lamborghini motor racing, Dilantha Malagamuwa won the Most Popular Sports Person of The Year Award.\newline  "BBC Sports Personality Team of the Year Award"\newline by Scottish teams; Celtic in 1967, after they became the first British football club to win the European Cup, and the 1990 Grand Slam winning Scotland rugby union squad. Football has had the highest representation among the winners, with 13 recipients. The most recent award was presented in 2017 to the England women's cricket team. This table lists the total number of awards won by nations that the teams have represented. This table lists the total number of awards won by the teams sporting discipline. BBC Sports Personality Team of the Year Award The BBC Sports Personality Team of the\newline "BBC Young Sports Personality of the Year"\newline In 2001, the award was replaced by the Young Sports Personality of the Year, and sprinter Amy Spencer was the first recipient of that award. Scottish tennis player Andy Murray, who won in 2004, is the only non-English recipient of the award. The only person to win the award more than once is diver Tom Daley, who won the award three times, in 2007, 2009, and 2010, and was nominated to the ten-person shortlist in five successive years (2007-2011). The most recent award was presented in 2017 to Manchester City midfielder Phil Foden. <nowiki>*</nowiki> Including a Newcomer of the Year\newline "Simone Biles"\newline Foundation. She was also one of the finalists for ""Time magazine""'s 2016 Person of the Year. Biles was also nominated for a 2016 ESPY award for Best Female Athlete along with Elena Delle Donne, Katie Ledecky, and Breanna Stewart; Stewart won the award. In July 2017, Biles won the ESPY Award for Best Female Athlete. She is the second gymnast to win this award after Nastia Liukin won it in 2009. In 2017, Simone won the Shorty Awards for the best in sports. At the 2017 Teen Choice Awards, Simone won favorite female athlete, and won Laureus World Sports Award\newline "BBC Sports Personality of the Year Award"\newline (2008) and Ennis-Hill (2017), received the BBC Sports Personality of the Year Lifetime Achievement Award. Princess Anne (1971) and her daughter Zara Phillips (2006) are the only award-winners to be members of the same family. The oldest recipient of the award is Dai Rees, who won in 1957 aged 44. Ian Black, who won the following year, aged 17, is the youngest winner. Torvill and Dean, who won in 1984, are the only non-individual winners of the award, so in the 61 years of the award there have been 62 recipients. Of these 13 have been female. 17 sporting disciplines
        \\ 
        \hline \hline
        \textbf{Without RAG answer:} Mo Farah \\
        \textbf{Top-1-passage RAG answer:} Greg Rusedski \\
        \textbf{Top-5-passage RAG answer:} Simone Biles \\

        \hline

    \end{tabular}
\end{table*}

\begin{table*}
    \centering
    \begin{tabular}{|p{2.0cm}||p{9cm}||p{1cm}||p{2.05cm}|}
        \hline
        \textbf{Method} & \textbf{Summary} & \textbf{words} & \textbf{Answer}\\
        \hline
        \hline
        RankGPT & "Newsfirst Platinum Awards"\newline Angelo Mathews won the Most Popular Sports Person of The Year Award. The second Platinum ceremony launching started with January 2017, with the full support of the Ministry of Education. Nominations for Platinum Awards 2017 opened under 20 categories. The promotional campaign commenced at Thurstan College, Colombo on 30 January 2017. The campaign was held in every district in the island until the 28 February 2017. The Awards night was held on 31 March 2017 at Sirasa Stein Studios, Ratmalana. Five times world champion in Lamborghini motor racing, Dilantha Malagamuwa won the Most Popular Sports Person of The Year Award. & 103 & Angelo Mathews
\\
\specialrule{0em}{2pt}{2pt}
        \scriptsize{\exper{LongLLMLingua}}& 0] "BBC Sports Personality of the Award\newline Document [3 "BBC Sports Person Team of the\newline Angelo Matws won The Pl ceremony launch started with17 with the support of the Ministry of.ations01 2ionalenced Th3 The campaign in every district in the the7 The night117asa Stein Studiosana world in motor racing, D Malwa of The Year Award.\newline Document [1] "BBC Young Sports Personality of the Year"\newline Document [4] "Simone Biles"\newline by Scottish teams; Celtic in 1967, after they became the first British football club to win the European Cup, and the 1990 Grand Slam winning Scotland rugby union squad. Football has had the highest representation among the winners, with 13 recipients. The most recent award was presented in 2017 to the England women's cricket team. This table lists the total number of awards won by nations that the teams have represented. This table lists the total number of awards won by the teams sporting discipline. BBC Sports Personality Team of the Year Award The BBC Sports Personality Team of the\newline who won the 2017 sports personality of the year
        &177 & Simone Biles
\\
\specialrule{0em}{2pt}{2pt}
        \exper{Llama2} & * Winner of the 2017 BBC Sports Personality of the Year: Simone Biles \newline * Winner of the 2017 BBC Sports Personality Team of the Year: England women's cricket team \newline * Winner of the 2017 Platinum Awards Most Popular Sports Person of the Year: Angelo Mathe & 45 & Simone Biles
\\
\specialrule{0em}{2pt}{2pt}
        \textbf{Ours} & BBC Sports Personality Team of the Year Award The BBC Sports Personality Team of the "Simone Biles" Foundation.\newline Contribution: [No] & 0 & Mo Farah \\
        \hline
        
        \hline
    \end{tabular}
    \caption{An example of empty compressed content on NQ. }
    \label{tab:nq_empty}
\end{table*}

\begin{table*}
    \centering
    \begin{tabular}{|p{15.6cm}|}
        \hline  
        
        \hline
        \textbf{Question}: who did the dominican republic gain its independence from \\
        \textbf{Standard Answer}: Haiti \\
        \hline \hline
        \textbf{Retrienval Passages} \\
        "Dominican Day Parade"\newline leadership of General Gregorio Luperon, the war was ultimately won from Spain. In 1844, the Dominican Republic secured its independence from Haiti and became a sovereign state until 1861. Under the leadership of General Pedro Santana, segments of the Dominican population sought to annex the Republic back to Spain and did so during March 18, 1861. On August 16, 1863, the start of the war for the Restoration of the Dominican Republic under the command of General Luperon. The Dominican Republic originally declared its independence from Spain on December 1, 1821. Ultimately, the Dominican Republic was re-established, free from Spain,\newline \newline "Colorism in the Caribbean"\newline into preference for lighter skin. Ritualistic skin bleaching to lighten one\u2019s skin, brown paper bag tests to verify one's skin tone, and degradation of darker-complected Haitians as ugly are contemporary manifestations of colorism in Haiti. After declaring its independence from Spanish rule in 1821, the Dominican Republic was overtaken by Haitian rule in 1822. The Dominican Republic did not achieve independence from Haiti until after their victory in the Dominican War of Independence in 1844. However, the country fell back under Spanish rule until it reclaimed its sovereignty after the Dominican War of Restoration of 1865. As the Dominican Republic\newline \newline "La Trinitaria (Dominican Republic)"\newline in August 1843 as a result of his dissident activities. La Trinitaria's other members continued the fight in Duarte's absence. One of them was Francisco del Rosario S\u00e1nchez, who corresponded with Duarte during the latter's exile in Venezuela, and Mat\u00edas Ram\u00f3n Mella, who along with Duarte and Sanchez became known as the founding fathers of the Dominican Republic. On January 16, 1844 La Trinitaria's manifesto in favor of independence was released, and the fight for independence began to gain the necessary momentum. On the heels of La Trinitaria's work, and after many battles and much bloodshed, the Dominican Republic was\newline \newline "Dominican Navy"\newline Dominican Navy The Navy of the Dominican Republic or Armada de Republica Dominicana (ARD), is one of the three branches of the Military of the Dominican Republic, together with the Army and the Air Force. After the Dominican Republic gained its independence from Haiti on February 27, 1844, there was a need to create a naval fleet. Three schooners were commissioned for this, (flagship), ""Mar\u00eda Chica"" and ""Leonor"". These were the original three Dominican vessels which were incorporated in the newly created Dominican Navy as authorized by the Junta Central Gubernativa with the Naval Act of 1844 on April 23,\newline \newline "Dominican Navy"\newline Brigade, which was under Spanish command and operating in Southern Iraq. Raised in 2008 the MIC serves as the youngest arm of service in the Dominican Navy, which is tasked for amphibious and landing operations. One Marine Battalion is of active service as of the present. It also the youngest Marine unit in all of Latin America. Dominican Navy The Navy of the Dominican Republic or Armada de Republica Dominicana (ARD), is one of the three branches of the Military of the Dominican Republic, together with the Army and the Air Force. After the Dominican Republic gained its independence from        
\\ 
        \hline \hline
        \textbf{Without RAG answer:} Spain \\
        \textbf{Top-1-passage RAG answer:} Spain \\
        \textbf{Top-5-passage RAG answer:} Haiti \\

        \hline

    \end{tabular}
\end{table*}

\begin{table*}
    \centering
    \begin{tabular}{|p{2.0cm}||p{9cm}||p{1cm}||p{2.05cm}|}
        \hline
        \textbf{Method} & \textbf{Summary} & \textbf{words} & \textbf{Answer}\\
        \hline
        \hline
        RankGPT & "Dominican Day Parade"\newline leadership of General Gregorio Luperon, the war was ultimately won from Spain. In 1844, the Dominican Republic secured its independence from Haiti and became a sovereign state until 1861. Under the leadership of General Pedro Santana, segments of the Dominican population sought to annex the Republic back to Spain and did so during March 18, 1861. On August 16, 1863, the start of the war for the Restoration of the Dominican Republic under the command of General Luperon. The Dominican Republic originally declared its independence from Spain on December 1, 1821. Ultimately, the Dominican Republic was re-established, free from Spain, & 103 & Spain
\\
\specialrule{0em}{2pt}{2pt}
        \scriptsize{\exper{LongLLMLingua}}& 0] "Domin Day Par\newline  [4 "La Trinitaria (Dominican\newline \newline Bade, was under Spanish command and operatingq. R0IC serves as theest arm of service in the Dominican, which task amphibious and operations. service of present also youngest in all of America.ican Navy The Navy of the Dominican Republic or Armada de), of Dominican Republic.ican Republic gained its\newline  [ "\newline ican Navy [Color the Caribbean"\newline \newline leadership of General Gregorio Luperon, the war was ultimately won from Spain. In 1844, the Dominican Republic secured its independence from Haiti and became a sovereign state until 1861. Under the leadership of General Pedro Santana, segments of the Dominican population sought to annex the Republic back to Spain and did so during March 18, 1861. On August 16, 1863, the start of the war for the Restoration of the Dominican Republic under the command of General Luperon. The Dominican Republic originally declared its independence from Spain on December 1, 1821. Ultimately, the Dominican Republic was re-established, free from Spain,\newline \newline who did the dominican republic gain its independence from & 171 & The Dominican Republic gained its independence from Spain.

\\
\specialrule{0em}{2pt}{2pt}
        \exper{Llama2} & The Dominican Republic gained its independence from Haiti on February 27, 1844, after the Dominican War of Independence. Prior to that, the country had gained independence from Spain in 1821, but was overtaken by Haitian rule in 1822. The Dominican R & 41 & Haiti
\\
\specialrule{0em}{2pt}{2pt}
        \textbf{Ours} & In 1844, the Dominican Republic secured its independence from Haiti and became a sovereign state until 1861. \newline Contribution: [Yes] & 17 & Haiti \\
        \hline
        
        \hline
    \end{tabular}
    \caption{An example of the most concise compressed content on NQ. }
    \label{tab:nq_concise}
\end{table*}

\begin{table*}
    \centering
    \begin{tabular}{|p{15.6cm}|}
        \hline  
        
        \hline
        \textbf{Question}: What was Walter Matthau's first movie? \\
        \textbf{Standard Answer}: The Kentuckian \\
        \hline \hline
        \textbf{Retrienval Passages} \\
        "Pete 'n' Tillie"\newline Walter Matthau received a Golden Globe nomination for Best Actor \u2013 Motion Picture Musical or Comedy, and won the 1973 BAFTA Award for Best Actor in a Leading Role for his performance in this movie and for his performance in ""Charley Varrick"". Carol Burnett received a Golden Globe Award nomination for Best Actress - Motion Picture Musical or Comedy. Pete 'n' Tillie Pete 'n' Tillie is a 1972 American comedy-drama film directed by Martin Ritt and starring Walter Matthau and Carol Burnett. Its advertising tagline was: ""Honeymoon's over. It's time to get married."" Screenwriter Julius J. Epstein was nominated for\newline \newline "Movers \& Shakers"\newline Movers \& Shakers Movers \& Shakers is a 1985 American comedy film distributed by MGM, starring Walter Matthau and directed by William Asher. The story follows the head of production at a Hollywood studio who wants to make a movie to fulfill a promise made to a dying friend. The film was written by Charles Grodin, who also appears in the movie. The cast includes Tyne Daly, Gilda Radner, and Vincent Gardenia. Steve Martin makes a cameo appearance as Fabio Longio. Hollywood studio mogul Joe Mulholland (Matthau) vows to produce the pet project of a dying acquaintance, who has been\newline \newline "The Fortune Cookie"\newline intention to sue the insurance company lawyers for invasion of privacy and report Purkey's racist remarks to various organizations. Hinkle drives to the stadium, where he finds Boom-Boom ready to leave the team and perhaps become a wrestler named ""The Dark Angel"". Hinkle manages to snap Boom-Boom out of his funk, and the two run down the fields passing and lateraling a football back and forth between them. This was the first film to feature the movie partnership of Jack Lemmon and Walter Matthau, who were to appear together in ten films: \u2020<br> Matthau and Lemmon also both had roles\newline \newline "The Secret Life of an American Wife"\newline a movie star who is known as an international sex symbol (Walter Matthau). Upon hearing that The Movie Star (the character is not given a name, and Matthau is credited as ""The Movie Star"" in the closing credits) indulges in the services of prostitutes, Victoria decides to pose surreptitiously as one in order to prove to herself that she is still sexually attractive. The Secret Life of an American Wife The Secret Life of an American Wife is a 1968 comedy film written and directed by George Axelrod. The film was released by 20th Century Fox in 1968, and was\newline \newline "Charles Matthau"\newline Charles Matthau Charles ""Charlie"" Matthau (born December 10, 1962) is a film and television director and actor and the son of actor Walter Matthau and actress/author Carol Saroyan. Born in New York City, he appeared as a child actor alongside his father in such films as ""Charley Varrick"" (1973), ""The Bad News Bears"" (1976) and ""House Calls"" (1978). Among his directorial projects have been ""The Grass Harp"", from a novella by Truman Capote, and the made-for-TV movie ""The Marriage Fool"", both of which starred his father. He also directed ""Doin' Time on Planet Earth"" (1988), ""Her Minor Thing"" (2005), ""Baby-O""

        \\ 
        \hline \hline
        \textbf{Without RAG answer:} The Kentuckian \\
        \textbf{Top-1-passage RAG answer:} The Taking of Pelham One Two Three \\
        \textbf{Top-5-passage RAG answer:} Pete 'n' Tillie \\

        \hline

    \end{tabular}
\end{table*}

\begin{table*}
    \centering
    \begin{tabular}{|p{2.0cm}||p{9cm}||p{1cm}||p{2.05cm}|}
        \hline
        \textbf{Method} & \textbf{Summary} & \textbf{words} & \textbf{Answer}\\
        \hline
        \hline
        RankGPT & "Pete 'n' Tillie"\newline Walter Matthau received a Golden Globe nomination for Best Actor \u2013 Motion Picture Musical or Comedy, and won the 1973 BAFTA Award for Best Actor in a Leading Role for his performance in this movie and for his performance in ""Charley Varrick"". Carol Burnett received a Golden Globe Award nomination for Best Actress - Motion Picture Musical or Comedy. Pete 'n' Tillie Pete 'n' Tillie is a 1972 American comedy-drama film directed by Martin Ritt and starring Walter Matthau and Carol Burnett. Its advertising tagline was: ""Honeymoon's over. It's time to get married."" Screenwriter Julius J. Epstein was nominated for & 103 & Pete 'n' Tillie
\\
\specialrule{0em}{2pt}{2pt}
        \scriptsize{\exper{LongLLMLingua}}& 1] "les Matthau\newline Document [Pete 'n' Till\newline  "The Fortune\newline \newline a movie star is as (W). that character nameauited Star) the, pose is. Theife 9 film Fox\newline ention and report Purkey's racist remarks to various organizations. Hinkle drives to the stadium, where he finds Boom-Boom ready to leave the team and perhaps become a wrestler named ""The Dark Angel"". Hinkle manages to snap Boom-Boom out of his funk, and the two run down the fields passing and lateraling a football back and forth between them. This was the first film to feature the movie partnership of Jack Lemmon and Walter Matthau, who were to appear together in ten films: \u2020<br> Matthau and Lemmon also both had roles\newline \newline What was Walter Matthau's first movie? & 122 & The Fortune

\\
\specialrule{0em}{2pt}{2pt}
        \exper{Llama2} & Walter Matthau's first movie was "The Secret Life of an American Wife" in 1968. & 14 & The Secret Life of an American Wife
\\
\specialrule{0em}{2pt}{2pt}
        \textbf{Ours} & "The Secret Life of an American Wife"\newline a movie star who is known as an international sex symbol (Walter Matthau). Carol Burnett received a Golden Globe Award nomination for Best Actress - Motion Picture Musical or Comedy. This was the first film to feature the movie partnership of Jack Lemmon and Walter Matthau, who were to appear together in ten films: \u2020<br> Matthau and Lemmon also both had roles\newline \"Movers \& Shakers\"\newline Movers \& Shakers Movers \& Shakers is a 1985 American comedy film distributed by MGM, starring Walter Matthau and directed by William Asher.\newline Contribution: [No] & 0 & The Kentuckian
 \\
        \hline
        
        \hline
    \end{tabular}
    \caption{An example of empty compressed content on TriviaQA. }
    \label{tab:tqa_empty}
\end{table*}

\begin{table*}
    \centering
    \begin{tabular}{|p{15.6cm}|}
        \hline  
        
        \hline
        \textbf{Question}: Which was the only eastern bloc country to participate in the 1984 LA Olympics? \\
        \textbf{Standard Answer}: Romania \\
        \hline \hline
        \textbf{Retrienval Passages} \\
        "1984 Summer Olympics boycott"\newline the majority of Soviet Bloc countries will not participate in the Games, Ceau\u0219escu's Romania is expected to attend. 1984 Summer Olympics boycott The boycott of the 1984 Summer Olympics in Los Angeles followed four years after the U.S.-led boycott of the 1980 Summer Olympics in Moscow. The boycott involved 14 Eastern Bloc countries and allies, led by the Soviet Union, which initiated the boycott on May 8, 1984. Boycotting countries organized another major event, called the Friendship Games, in July and August 1984. Although the boycott led by the Soviet Union affected a number of Olympic events that were normally\newline \newline "Nicolae Ceaus\u0326escu"\newline visit of Egyptian president Anwar Sadat to Israel in 1977. Also Romania was the only country in the world to maintain normal diplomatic relations with both Israel and the PLO. In 1980, Romania participated in the 1980 Summer Olympics in Moscow with its other Soviet bloc allies, but in 1984 was one of the few Communist countries to participate in the 1984 Summer Olympics in Los Angeles when most of the Eastern Bloc's nations boycotted this event. In 1966, Ceau\u0219escu, in an attempt to boost the country's population, made abortion illegal and introduced Decree 770 to reverse the low birth\newline \newline "Summer Olympic Games"\newline Eastern Bloc that did attend the 1984 Olympics. These games were perhaps the first games of a new era to make a profit. Although a boycott led by the Soviet Union depleted the field in certain sports, 140 National Olympic Committees took part, which was a record at the time. Again, without the participation of the Eastern European countries, the 1984 Games were dominated by their host country. The Games were also the first time mainland China (People's Republic) participated. According to British journalist Andrew Jennings, a KGB colonel stated that the agency's officers had posed as anti-doping authorities from\newline \newline "Romania at the Olympics"\newline Romania at the Olympics Romania first participated at the Olympic Games in 1900, with a single participant. The National Olympic Committee for Romania is the Romanian Olympic and Sports Committee, and was created and recognized in 1914. The nation first sent a team to compete at the Games in 1924, and has only missed two editions each of the Summer Olympic Games and Winter Olympic Games since then. Notably, Romania was the lone Eastern Bloc nation to participate at the 1984 Summer Olympics, which the other nations boycotted. That was also Romania's most successful Olympic Games: they won 20 gold\newline \newline "Craig Beardsley"\newline medal in the 200-meter butterfly at the Pan American Games in Caracas, Venezuela. Beardsley failed to qualify for the U.S Olympic team in 1984, by placing third by 0.36 of a second in the U.S. Olympic Trials, and he retired from competitive swimming afterward. Taking advantage of the hard feelings created by the American-led boycott of the 1980 Moscow Olympics, the Soviet Union and other Eastern Bloc countries had pushed for and received Olympic rules changes limiting each country to only two participants in each swimming event. The United States, as the dominant swimming power, was hurt the most. Beardsley
       
\\ 
        \hline \hline
        \textbf{Without RAG answer:} East Germany \\
        \textbf{Top-1-passage RAG answer:} Romania \\
        \textbf{Top-5-passage RAG answer:} Romania \\

        \hline

    \end{tabular}
\end{table*}

\begin{table*}
    \centering
    \begin{tabular}{|p{2.0cm}||p{9cm}||p{1cm}||p{2.05cm}|}
        \hline
        \textbf{Method} & \textbf{Summary} & \textbf{words} & \textbf{Answer}\\
        \hline
        \hline
        RankGPT & "1984 Summer Olympics boycott"\newline the majority of Soviet Bloc countries will not participate in the Games, Ceau\u0219escu's Romania is expected to attend. 1984 Summer Olympics boycott The boycott of the 1984 Summer Olympics in Los Angeles followed four years after the U.S.-led boycott of the 1980 Summer Olympics in Moscow. The boycott involved 14 Eastern Bloc countries and allies, led by the Soviet Union, which initiated the boycott on May 8, 1984. Boycotting countries organized another major event, called the Friendship Games, in July and August 1984. Although the boycott led by the Soviet Union affected a number of Olympic events that were normally & 104 & Romania
\\
\specialrule{0em}{2pt}{2pt}
        \scriptsize{\exper{LongLLMLingua}} & Document [1] "1984 Summer Olympics boycott\newline \newline Document [] "Sum Olympic Games"\newline  [ "Romania\newline \newline Eloc the4 These were first of a led the field was., the Eastern European4 by The also ( Republic participated.GB thes had- from will not participate in the Games, Ceau\u0219escu's Romania is expected to attend. 1984 Summer Olympics boycott The boycott of the 1984 Summer Olympics in Los Angeles followed four years after the U.S.-led boycott of the 1980 Summer Olympics in Moscow. The boycott involved 14 Eastern Bloc countries and allies, led by the Soviet Union, which initiated the boycott on May 8, 1984. Boycotting countries organized another major event, called the Friendship Games, in July and August 1984. Although the boycott led by the Soviet Union affected a number of Olympic events that were normally\newline \newline Which was the only eastern bloc country to participate in the 1984 LA Olympics? & 144 & Romania
\\
\specialrule{0em}{2pt}{2pt}
        \exper{Llama2} & Romania was the only Eastern Bloc country to participate in the 1984 Los Angeles Olympics, and it was also the country's most successful Olympic Games, winning 20 gold medals. & 29 & Romania
\\
\specialrule{0em}{2pt}{2pt}
        \textbf{Ours} & Notably, Romania was the lone Eastern Bloc nation to participate at the 1984 Summer Olympics, which the other nations boycotted. \newline Contribution: [Yes]& 20 & Romania
\\
        \hline
        
        \hline
    \end{tabular}
    \caption{An example of the most concise compressed content on TriviaQA. }
    \label{tab:tqa_concise}
\end{table*}

\begin{table*}
    \centering
    \begin{tabular}{|p{15.6cm}|}
        \hline  
        
        \hline
        \textbf{Question}: What is the year of the event that occured first, Making Today a Perfect Day was produced, or Frozen was produced? \\
        \textbf{Standard Answer}: 2013 \\
        \hline \hline
        \textbf{Retrienval Passages} \\
        "Making Today a Perfect Day"\newline Making Today a Perfect Day ""Making Today a Perfect Day"" is a song from the 2015 Walt Disney Animation Studios computer-animated short film ""Frozen Fever"", with music and lyrics by Kristen Anderson-Lopez and Robert Lopez and performed throughout most of the short. It was released as a single in the United States on March 12, 2015. On September 2, 2014, during the ABC airing of """", Walt Disney Animation Studios' chief creative officer John Lasseter announced that a ""Frozen"" short film with a new song would be released in the future. On the same day, ""Variety"" announced that the short\newline \newline "Making Today a Perfect Day"\newline the lyrics off-by-heart. Kat Brown of ""The Daily Telegraph"" referred to the short film as a ""musical video"", due to such a large proportion of it being taken up by this song. ""Us Weekly"" negatively compared its catchiness to ""Let It Go"", though described the ditty as ""fresh"", ""bright"", and ""fun"". In a negative review, ""Slate"" felt that ""the song itself, while hummable, is fatally damaged by its need to do too much."" Making Today a Perfect Day ""Making Today a Perfect Day"" is a song from the 2015 Walt Disney Animation Studios computer-animated short film ""Frozen Fever"", with music\newline \newline "Making Today a Perfect Day"\newline would be released in early 2015 under the title ""Frozen Fever"", with Chris Buck and Jennifer Lee returning as co-directors, Peter Del Vecho returning as producer and a new song by Kristen Anderson-Lopez and Robert Lopez. In a mid-October interview, Idina Menzel revealed that the cast had already recorded their vocal tracks, stating ""We just worked on a short for ""Frozen""."" On December 3, 2014, it was announced that Aimee Scribner would be a co-producer and that ""Frozen Fever"" would debut in theaters alongside Walt Disney Pictures' ""Cinderella"" on March 13, 2015. In late December, the co-directors told the Associated\newline \newline "Making Today a Perfect Day"\newline to start brainstorming possibilities. After early discussions about Olaf, head story artist Marc Smith pitched the idea of what might happen if Elsa had a cold, which became the basis for the short's plot. The directors began working on the short in June and by August were back in the recording studio with the cast to lay down vocal tracks. The short features the song ""Making Today a Perfect Day"", by Anderson-Lopez and Lopez. At the premiere of ""Cinderella"" and ""Frozen Fever"" at the El Capitan Theatre in Hollywood, California, on March 1, 2015, Josh Gad told ""USA Today"", ""I\newline \newline "Making Today a Perfect Day"\newline want to apologize to parents everywhere for the fact that children are going to be singing a whole new ""Frozen"" song..."" Gad's wife noticed he was still humming it two days after he recorded his lines. In ""Making Today a Perfect Day"" there is a lyrical reference to ""Let It Go""\u2014Elsa notes to Anna that ""a cold never bothered me anyway"", this time re-purposing the line by referring to an actual cold. ""Billboard"" suggests that the songwriting duo included this Easter egg because they ""know exactly what the fanbase wanted"". The beginning of the song also includes a passage to

        \\ 
        \hline \hline
        \textbf{Without RAG answer:} 2013 \\
        \textbf{Top-1-passage RAG answer:} 2015 \\
        \textbf{Top-5-passage RAG answer:} Frozen was produced first. \\

        \hline

    \end{tabular}
\end{table*}

\begin{table*}
    \centering
    \begin{tabular}{|p{2.0cm}||p{9cm}||p{1cm}||p{2.05cm}|}
        \hline
        \textbf{Method} & \textbf{Summary} & \textbf{words} & \textbf{Answer}\\
        \hline
        \hline
        RankGPT & "Making Today a Perfect Day"\newline Making Today a Perfect Day ""Making Today a Perfect Day"" is a song from the 2015 Walt Disney Animation Studios computer-animated short film ""Frozen Fever"", with music and lyrics by Kristen Anderson-Lopez and Robert Lopez and performed throughout most of the short. It was released as a single in the United States on March 12, 2015. On September 2, 2014, during the ABC airing of """", Walt Disney Animation Studios' chief creative officer John Lasseter announced that a ""Frozen"" short film with a new song would be released in the future. On the same day, ""Variety"" announced that the short & 105 & 2015
\\
\specialrule{0em}{2pt}{2pt}
        \scriptsize{\exper{LongLLMLingua}}& Making Today a Perfect ""aking Today a Perfect Day is a song from the 20 Walt Disney Animation Studios-animated short ""Fro F"", music and lyrics Kristen Anderson-Lopez and Robert Lopez throughout most of short It was single in United States on March 12, 2015. On September 2, 2014, during ABC airing of Walt Disney Animation chief creative officer John Lass announced aFrozen"" short film with a new would be released future same day,Vari announced that the\newline  [Making Today a\newline \newline Document [4M Day [fect Day"\newline Frozen Fever"", with music\newline \newline What is the year of the event that occured first, Making Today a Perfect Day was produced, or Frozen was produced? & 109 & 2015
\\
\specialrule{0em}{2pt}{2pt}
        \exper{Llama2} & The event "Making Today a Perfect Day" was produced first, in 2015. & 12 & 2015
\\
\specialrule{0em}{2pt}{2pt}
        \textbf{Ours} & The short features the song ""Making Today a Perfect Day"", by Anderson-Lopez and Lopez.\newline Contribution: [No] & 0 & 2013
\\
        \hline
        
        \hline
    \end{tabular}
    \caption{An example of empty compressed content on HotpotQA. }
    \label{tab:hotpot_empty}
\end{table*}

\begin{table*}
    \centering
    \begin{tabular}{|p{15.6cm}|}
        \hline  
        
        \hline
        \textbf{Question}: Salisbury Woodland Gardens links a zoo with a park designed and built under the watchful eye of who? \\
        \textbf{Standard Answer}: Thomas Mawson \\
        \hline \hline
        \textbf{Retrienval Passages} \\
        "Salisbury Woodland Gardens, Blackpool"\newline enable the local community to get more involved in the sites management and interpretation. Salisbury Woodland Gardens, Blackpool Salisbury Woodland Gardens is an open space located in the east of Blackpool, flanked by East Park Drive and Woodside Drive and linking Blackpool Zoo with Stanley Park. Known simply as the 'Woodland Gardens' to local people, the site was acquired in 1924 by Blackpool Corporation and was originally developed as a shelter belt for the adjacent Stanley Park Golf Course. The gardens were later developed in the 1940s as an arboretum and public open space for all to enjoy. It was\newline \newline "Salisbury Woodland Gardens, Blackpool"\newline Salisbury Woodland Gardens, Blackpool Salisbury Woodland Gardens is an open space located in the east of Blackpool, flanked by East Park Drive and Woodside Drive and linking Blackpool Zoo with Stanley Park. Known simply as the 'Woodland Gardens' to local people, the site was acquired in 1924 by Blackpool Corporation and was originally developed as a shelter belt for the adjacent Stanley Park Golf Course. The gardens were later developed in the 1940s as an arboretum and public open space for all to enjoy. It was renovated in 1967 by Peter Perry and his 'Flying Squad (see below). Popular once\newline \newline "Salisbury Woodland Gardens, Blackpool"\newline as a wedding photograph location, the site went into decline during the 1990s. The Council's Ranger Service manage and protect the gardens which they took over in September 2006 and have been funding and undertaking the restoration of the woodland. In 1967, Parks Director Norman Leach appointed gardener Pete Perry and his Flying Squad of gardeners to plant up the gardens. All plants, (primulas, meconopsis, etc.) were grown from seed in the greenhouses at Stanley Park, and planted ""en masse"". Extra shrubs, including azalea were also planted. The neighbouring Blackpool Zoo site was formerly Blackpool's municipal airport. In 1927 the\newline \newline "Stanley Park, Blackpool"\newline Stanley Park, Blackpool Stanley Park is a public park in the town of Blackpool on the Fylde coast in Lancashire, England. It is the town's primary park and covers an area of approximately . The park was designed to include significant sporting provisions, along with formal gardens, a boating lake and woodland area. It was designed and built in the 1920s, under the eye of Thomas Mawson. It is located in the Great Marton and Layton areas of the town. It is Grade II* listed and is on the Register of Historic Parks and Gardens of special historic interest in\newline \newline "ZooTampa at Lowry Park"\newline Larry Killmar, the zoo's Director of Collections who had authorized many of Salisbury's questionable animal transfers. Under Killmar, the zoo reorganized its internal policies over several months, and on March 27, 2009, the AZA reinstated the membership of both Lowry Park Zoo and its director of collections. The saga came to a close in August 2009 when Salisbury and the Lowry Park Zoo board agreed to a settlement in which Salisbury paid \$2,200 and agreed to return all the structures, fencing, and equipment that the zoo had built at Safari Wild but did not admit to any wrongdoing. ZooTampa at
   
\\ 
        \hline \hline
        \textbf{Without RAG answer:} Capability Brown \\
        \textbf{Top-1-passage RAG answer:} Blackpool Corporation \\
        \textbf{Top-5-passage RAG answer:} Peter Perry. \\

        \hline

    \end{tabular}
\end{table*}

\begin{table*}
    \centering
    \begin{tabular}{|p{2.0cm}||p{9cm}||p{1cm}||p{2.05cm}|}
        \hline
        \textbf{Method} & \textbf{Summary} & \textbf{words} & \textbf{Answer}\\
        \hline
        \hline
        RankGPT & "Salisbury Woodland Gardens, Blackpool"\newline enable the local community to get more involved in the sites management and interpretation. Salisbury Woodland Gardens, Blackpool Salisbury Woodland Gardens is an open space located in the east of Blackpool, flanked by East Park Drive and Woodside Drive and linking Blackpool Zoo with Stanley Park. Known simply as the 'Woodland Gardens' to local people, the site was acquired in 1924 by Blackpool Corporation and was originally developed as a shelter belt for the adjacent Stanley Park Golf Course. The gardens were later developed in the 1940s as an arboretum and public open space for all to enjoy. It was & 104 & Blackpool Corporation
\\
\specialrule{0em}{2pt}{2pt}
        \scriptsize{\exper{LongLLMLingua}}& Document [] "Salisbury Woodland Gard Blackpool\newline \newline  [1 "Salisbury Woodland Gard, Blackpool"\newline ]Sbury,\newline \newline Sburypool located eastpool by East Drive Drive and linkingpooloo withown asood local site acquired4pool Corporationt for Theens were0um and for. It97 by andFlying Squad (see below). Popular once\newline \newline enable the local community to get more involved in the sites management and interpretation. Salisbury Woodland Gardens, Blackpool Salisbury Woodland Gardens is an open space located in the east of Blackpool, flanked by East Park Drive and Woodside Drive and linking Blackpool Zoo with Stanley Park. Known simply as the 'Woodland Gardens' to local people, the site was acquired in 1924 by Blackpool Corporation and was originally developed as a shelter belt for the adjacent Stanley Park Golf Course. The gardens were later developed in the 1940s as an arboretum and public open space for all to enjoy. It was\newline \newline Salisbury Woodland Gardens links a zoo with a park designed and built under the watchful eye of who? & 158 & Flying Squad
\\
\specialrule{0em}{2pt}{2pt}
        \exper{Llama2} & Salisbury Woodland Gardens in Blackpool was designed and built under the watchful eye of Thomas Mawson in the 1920s, as part of Stanley Park. The gardens were later renovated in 1967 by Pete Perry and his 'Flying Squad'. The site is now managed by th & 45 & Thomas Mawson
\\
\specialrule{0em}{2pt}{2pt}
        \textbf{Ours} & It was designed and built in the 1920s, under the eye of Thomas Mawson. It was designed and built in the 1920s, under the eye of Thomas \newline Contribution: [Yes]& 27 & Thomas Mawson
\\
        \hline
        
        \hline
    \end{tabular}
    \caption{An example of the most concise compressed content on HotpotQA. }
    \label{tab:hotpot_concise}
\end{table*}

\end{document}